\documentclass[runningheads]{llncs}
\usepackage{times}
\usepackage{graphicx}
\usepackage{latexsym}
\usepackage{subcaption}
\usepackage{amsmath}

\usepackage{hyperref}
\usepackage{url}
\usepackage{csquotes}
\usepackage{float}
\usepackage{array,multirow,graphicx}
\usepackage{xcolor}

\usepackage{ntheorem}
\theoremseparator{:}

\newtheorem{nullhypothesis}{Null Hypothesis}
\begin{document}
\title{Birds of a Feather Flock Together: \\Satirical News Detection via \\Language Model Differentiation}
%
%
\author{Yigeng Zhang\inst{1} \and
Fan Yang\inst{1} \and
Yifan Zhang\inst{1}\and
Eduard Dragut\inst{2}\\and
Arjun Mukherjee\inst{1}}
\authorrunning{Y. Zhang et al.}
%
\institute{University of Houston, Houston, TX 77004, USA 
\email{\{yzhang168,fyang11,yzhang114\}@uh.edu, arjun@cs.uh.edu}\\
\and
Temple University, Philadelphia, PA 19122, USA\\
\email{edragut@temple.edu}}
\maketitle              
\vspace{-10pt}
\begin{abstract}

Satirical news is regularly shared in modern social media because it is entertaining with smartly embedded humor. However, it can be harmful to society because it can sometimes be mistaken as factual news, due to its deceptive character. We found that in satirical news, the lexical and pragmatical attributes of the context are the key factors in amusing the readers. In this work, we propose a method that differentiates the satirical news and true news. It takes advantage of satirical writing evidence by leveraging the difference between the prediction loss of two language models, one trained on true news and the other on satirical news, when given a new news article. We compute several statistical metrics of language model prediction loss as features, which are then used to conduct downstream classification. The proposed method is computationally effective because the language models capture the language usage differences between satirical news documents and traditional news documents, and are sensitive when applied to documents outside their domains. 

\keywords{Satirical news detection \and Text classification \and Deception detection.}
\end{abstract}
\section{Introduction} \label{Introduction}
Satirical news is a kind of literary work that consists of parodies of mainstream journalism, mundane events, or other humor. In the modern world, satirical news can be harmful to a society because it is deceptive in nature and hard to distinguish on many occasions. The creative approach of satire is witty, metaphorical, and subtle and people without a related cultural or contextual background may have difficulty in telling it apart from factual news items. Satirical news may have unintentional consequences similar to fake news \cite{rubin2016fake} and, thus, investigating the methods of filtering satirical news has drawn people's attention.

In recent years, there have been a surge of works on fake news detection; however, the aim of producing news satire is not to contradict the truth and misleading people as fake news does, but rather to entertain as form of parody. Satirical news articles have these characteristics:
\begin{itemize}
\item {\bf Imaginative  Content}: Similar to fake news, satirical news also entails fictional content \cite{rubin2016fake}. Fake news, pretending to report a real story, is intended to deliver false information to mislead people. The fake stories are created seemingly reasonable, thus people who are fooled will take it as fact without being skeptical. Although satirical news is also fictional, the purpose of creating satirical fake content is to make the readers aware of an irony and the humor behind it.
\item {\bf Seemingly Formal and Serious Writing Style}: Satirical news is usually written in a formal form and subjective tone in the same way as true news. Yang et al. reported that news satire is written in subjective tones, suggesting a formal form and mimicking true news \cite{yang2017satirical}. This makes satire hold a kind of humor by contrasting the serious writing style and ridiculous story. 
\item {\bf Contradicting Common Sense}: Once a reader deciphers the irony and deadpan humor in a satire article, the reader will realize its ridiculousness since the content violates common sense. Satirical news story sometimes combines irrelevant subjects together to create unexpectedness and humor \cite{reyes2012humor}. For example:
\begin{displayquote}
\emph{`Father spends joyful afternoon throwing son around backyard.'} \textemdash Onion
\end{displayquote} This sentence gives a sense of ridiculousness because the object `son' is not used to be `thrown' for fun around the backyard.
Sometimes the stories are made up of impossible events that will never happen in the real world if one knows the context. For example: 
\begin{displayquote}
\emph{`Vice President Mike Pence reportedly visited his conversion therapist Thursday for a routine gay-preventative checkup.'} \textemdash Onion
\end{displayquote} 
\item {\bf Humoristic and Amusing}: The purpose of creating news satire is to criticize or comment on some social affair in a humorous way \cite{burfoot2009automatic}. The readers usually find funny metaphors and comical stories in it. This entertaining property increases the popularity of news satire in social media outlets.
\end{itemize}

Although this seemingly formal writing style makes satirical news hard to detect, the breach of lexical and pragmatical information can address the problem.
Since satirical news usually makes up stories with preposterous content, it is easy for people with corresponding knowledge and cultural background to recognize it. Using a similar idea as how people discern illogical information, we can leverage the computational models which have the ability to gain domain knowledge to `judge' like human beings. Language models (LM) is one possible solution because an LM encodes knowledge from the context it is trained upon \cite{trinh2018simple}.
In this work, by making use of the characteristics of the satirical content, we propose a new method to detect satirical news, where language models play an important part. 

\section{Related Work}\label{RelatedWork}
In recent years, there have been many works in news and sentiment analysis \cite{DBLP:journals/widm/HeHMOD20}\cite{StanojevicADO19} \cite{SchneiderD15}\cite{DragutYSM10}. One particular area has been fake and satirical news detection using machine learning methods. Burfoot \& Baldwin conducted a research on differentiating satire news and true news using SVM with targeted lexical features and semantic validity on a 4000-article dataset \cite{burfoot2009automatic}. Rubin et al. categorized news satire as a kind of `humorous fakes' in deceptive news \cite{rubin2015deception}. They proposed an SVM-based algorithm in satire detecting and tested with 360 pieces of news \cite{rubin2016fake}. Yang et al. built a dataset for satirical news detection \cite{yang2017satirical}, which contains a much larger number of satirical and true news from various sources. They also proposed a method with Hierarchical Attention Networks with many linguistic features such as writing stylistic features and readability features. They argued that satirical cues often appear in certain paragraphs instead of the whole article. De Sarkar et al. works at the sentence level embeddings and document level embeddings, and they also use many syntax features such as part-of-speech tags and named entity features \cite{de2018attending}.

Compared to the existed satirical news detection works, we propose an approach that seeks to capture the characteristics mentioned in Section \ref{Introduction} about news satire and use it to distinguish it from other news genre. 
Our method utilizes two separately trained language models from satirical news and true news as the `brain' of domain knowledge. Then we obtain a series of satire/non-satire measurement scores\textemdash surprise scores for each news article from the language models. With these scores as features, a classifier is trained for future prediction. As the idiom `Birds of a feather flock together' goes, we find that news of the same category will have similar feature representations while news from different categories will show a significant difference when viewed through their corresponding language models.
The dataset we use for satirical news detection is from the work by Yang et al. \cite{yang2017satirical}.  Experimental results on this dataset show that our method can achieve state-of-the-art performance on the validation dataset and competitive results on the test data. Moreover, it only uses classical neural language models with shallow layers and a small number of features from several basic statistics of the language model output instead of sophisticated feature engineering.

\section{Method}\label{Method}
In this section, we first present the underlying hypothesis of our approach. Then we present our model and classification pipeline.
\subsection{Hypothesis}
Satirical news is written in a seemingly formal and serious way just like true news. But people can distinguish satirical news from true news because its content `violates people's common sense'. As discussed in Section \ref{Introduction}, satirical content contains stories which can hardly happen in real life. Looking at this phenomenon at the language level, the pairing of subject-object and the word collocations in a large number of satirical news articles appears to be significantly different from a true news collection. Since language models (LM) encodes knowledge from the data they are trained from, they are expected to act differently (i.e., present different scores) when fed with uncommon text.
We expect this to be reflected in entropy (logarithm of perplexity) when a pre-trained true news LM is used to fit true news from satire news. 

We assume that, because of the lexical and pragmatical differences, when true/satire news is applied to a pre-trained language model, news samples from a different category will result in an obvious difference as judged based on the output of the language model. By applying the Wilcoxon signed-rank test onto the output pairs from different language models, we expect to prove the result of significant difference \cite{mcdonald2009handbook}.

Here we define a metric named \emph{surprise score}.
A \emph{surprise score} is the arithmetic mean of the entropy loss values on token-level that the LM produces when fed in with a piece of sequential text. The more distinct the new text piece (from the training data) is, the higher the surprise score obtained with the LM we expect to be. For example, a piece of satirical news will have a higher surprise score than a true news after being applied to a language model trained on true news documents. By leveraging this surprise scores as features, we can perform the classification effectively. 
 

\subsection{Word-Level Neural Language Model with LSTM}
Language model (LM) can be defined as a probability distribution over a sequence of words. 
It is usually trained to describe the likelihood of occurrence of one next word $w_{t}$ in a sequence by having seen the previous $k$ words of the context:
\begin{equation}
p \left(w_ {t}|w_{t-k-1: t-1} \right) = LM\left(w_{t-k-1: t-1} \right) 
\end{equation}
where $\forall w_{t} \in V$ , the finite vocabulary of the context.

\begin{figure}[t]
  \centering
  \begin{minipage}[t]{0.5\textwidth}
    \includegraphics[width=\textwidth]{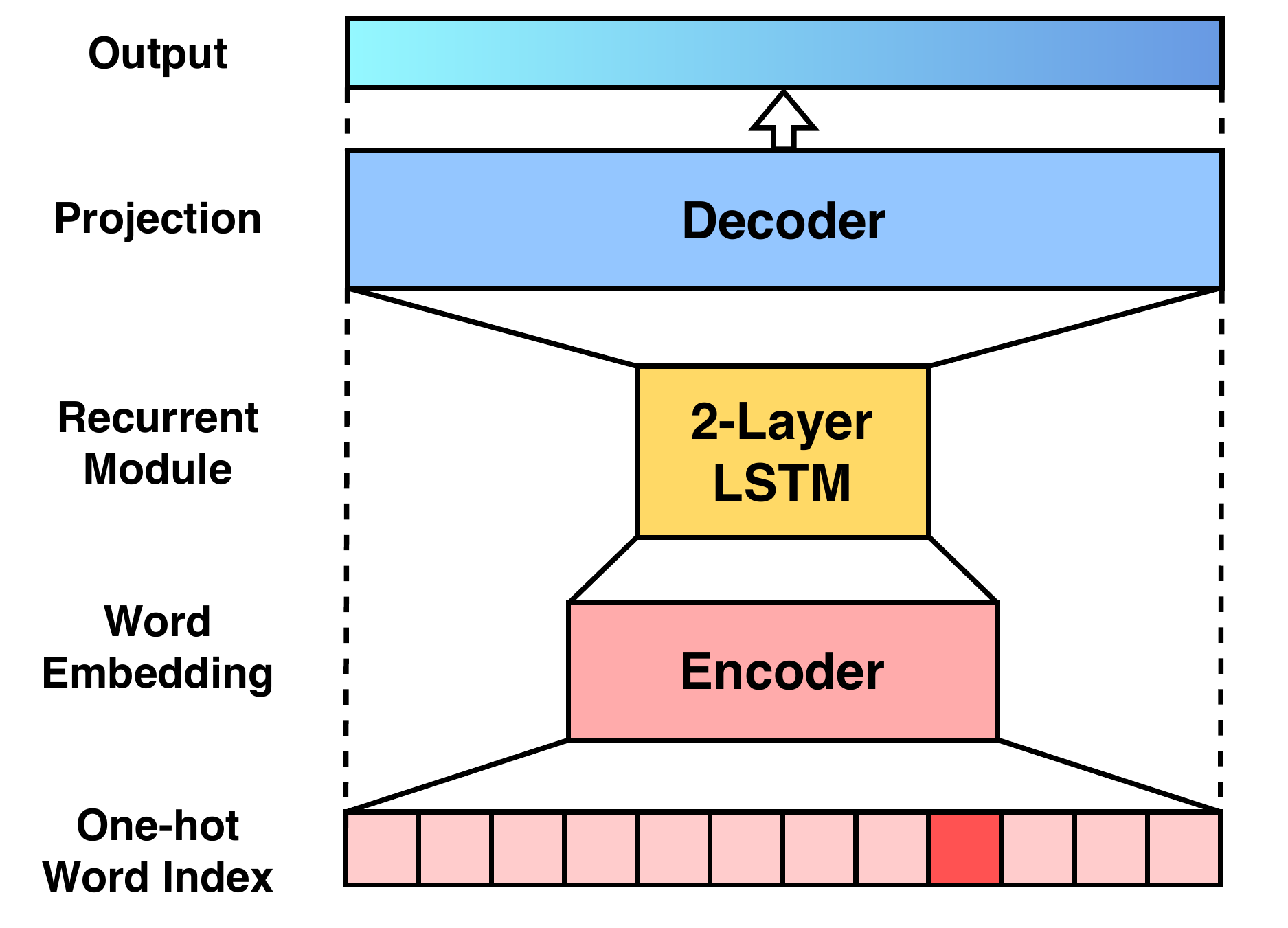}
    \caption{Basic encoder-decoder LM.}
    \label{fig:LM}
  \end{minipage}
    \hspace*{\fill}
  \begin{minipage}[t]{0.4\textwidth}
    \includegraphics[width=\textwidth]{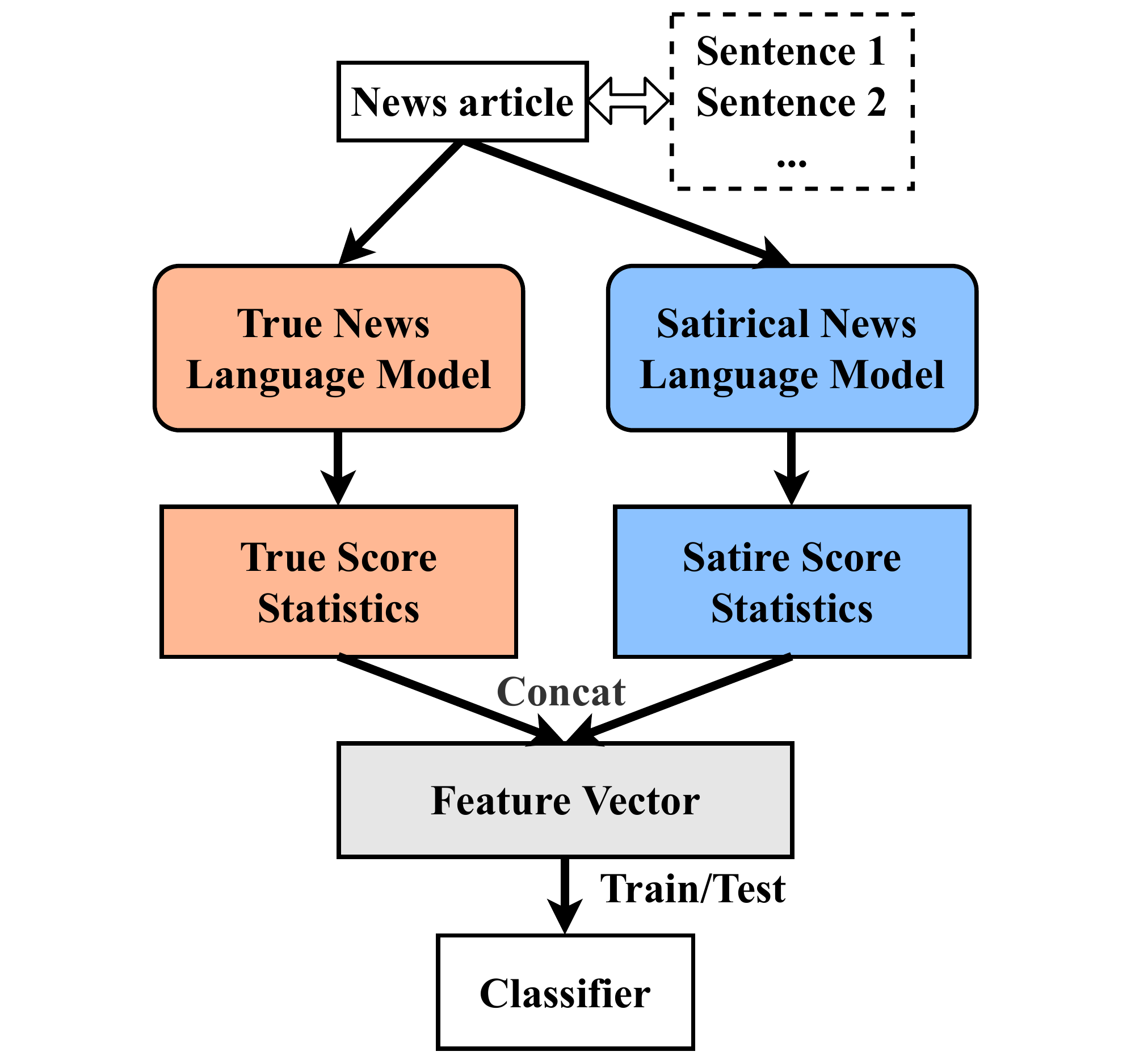}
    \caption{Pipeline.}
    \label{fig:pipeline}
  \end{minipage}
  \vspace{-20pt}
\end{figure}

The recurrent neural network based language model (RNN LM) was firstly proposed by  Mikolov et al. \cite{mikolov2010recurrent}. The neural language model used in this work follows a basic encoder-decoder language model with LSTM as the recurrent module (shown in Figure \ref{fig:LM}).
The prediction procedure is derived by:
\begin{eqnarray}
w _ { t }  = L y _ { t - 1 } ^ { * }
\\
h _ { t }  = f \left( w _ { t } , h _ { t - 1 } \right) 
\\
y _ { t }  =  W h _ { t } + b
\end{eqnarray}

The matrix $L \in  { R } ^ { d _ { x } \times | V | }$ is for word embedding. $f$ refers to the LSTM module. The $W$ is the $w_{t}$ matrix. After linear transformation from the decoder, output $y _ { t }$ is obtained. The cross-entropy loss on sequence is calculated as below. In this loss function, $x$ is a raw output score vector from the linear projection layer for each class, and $i$ is the dictionary index of each corresponding word, which indicates the class index number.
\vspace{-10pt}

\begin{eqnarray}
\mathcal{L}(\mathbf{{y}^{target}}, \mathbf{\hat{y}}) = -\sum_{i=1}^{V} y_{i}^{target} \log \hat{y}_{i}
\end{eqnarray}
\vspace{-20pt}

\subsection{Pipeline}
We favor a less complex classification pipeline in this work. The input news article is fed as a word sequence into the two language models, which are each trained on true news and satirical news documents, respectively. From each language model, we get an output sequence of loss numbers corresponding to each sentence of one article. Now each article of $n$ sentences is represented with two corresponding sequences, describing the impression of true and satirical news language models:
\begin{align*} 
Article_{i}^{true} = {score_{1}^{t},score_{2}^{t},score_{3}^{t},...,score_{n}^{t}}
\\
Article_{i}^{satire} = {score_{1}^{s},score_{2}^{s},score_{3}^{s},...,score_{n}^{s}}
\end{align*}
   
We calculate the mathematical statistics: \textbf{sample size} $N$, \textbf{arithmetic mean} $\bar{X}$, \textbf{median} $\tilde{X}$, \textbf{sample variance} $s^2$, and \textbf{range} $R$, of each sequence of surprise scores as satire/true features, where \textbf{median} is the middle value and \textbf{sample size} represents the number of sentences of each news article. Then we concatenate all of the features as an 9-dimension vector to represent one news article. For each article, we finally obtained a low-dimension feature vector:
\begin{equation}
 [ N, \bar{X_t}, \tilde{X_t}, s^2_t,{R_t},
 \bar{X_s}, \tilde{X_s}, s^2_s,{R_s}]
\end{equation}
where the subscript $s$ (for satirical) or $t$ (for true) indicates which language model the corresponding feature comes from. An SVM classifier is trained and tested using the above feature vectors. Figure \ref{fig:pipeline} shows the proposed classification pipeline.

\section{Experiment and Evaluation}\label{ExperimentandEvaluation}
In this section, we first present the dataset and hypothesis testing. Then we introduce the implementation details. Finally, we dive into the evaluation and analysis.
\subsection{Dataset}
The news dataset we use in this paper is from the work by Yang et al. \cite{yang2017satirical}. The news articles are crawled from both satirical news providers (such as Onion) and true news websites (such as CNN). The news headline, creation time, and author information are removed in order not to introduce many obvious features of a news source. 

In this work, we concentrate on the binary classification task. The usage of this news dataset could be found in Table \ref{tab:data}. We divide the original training data from \cite{yang2017satirical} into two parts: the part to train the two language models and the part to train the classifier. The former part takes 2/3 of the original training data, while the latter takes the rest 1/3. The data for validation and test remains the same.
\begin{table}[t]
\centering
\caption{The usage distribution of the News dataset.}
\label{tab:data}
\resizebox{0.75\textwidth}{!}{%
\begin{tabular}{|l|l|l|l|l|l|}
\hline
       & Original Train & \textbf{Train LM} & \textbf{Train SVM} & Validation & Test  \\ \hline
True   & 101268         & \textbf{67512}    & \textbf{33756}     & 33756      & 33756 \\ \hline
Satire & 9538           & \textbf{6358}     & \textbf{3180}      & 3103       & 3608  \\ \hline
\end{tabular}%
}
\vspace{-10pt}
\end{table}
The part of data for training the classifier is roughly the same size as the validation data and test data. This practice of division is to ensure the balance of each part of data usage and also to prove the effectiveness and generalization ability of the classifier. 
Moreover, this dataset is from various news sources -- Train: \emph{Onion}, \emph{the Spoof}; Validation: \emph{Daily Current}, \emph{Daily Report}, \emph{Enduring Vision}, \emph{Gomerblog}, \emph{National Report}, \emph{Satire Tribune}, \emph{Satire Wire}, \emph{Syruptrap}, and \emph{Unconfirmed Source}; Test: \emph{Satire World}, \emph{Beaverton} and \emph{Ossurworld}, which makes the data distribution and its characteristics not uniform. In this case, the adaptability of the proposed method will be tested since the language models and the classifier are trained on different news sources. Therefore, because of the diversity of the news sources in this dataset, it can finally help to disclose whether our method has the ability to generally catch the knowledge difference behind satirical content and true content.



\subsection{Statistical Hypothesis Testing}
As mentioned in section \ref{Method}, we should prove that the statistics from true/satire surprise scores of each given article produced from both true and satire LMs have a significant difference. Therefore, we need statistical hypothesis testing to examine the score pairs are deemed statistically significant. 

\setcounter{nullhypothesis}{-1}
\begin{nullhypothesis}[$H_{\ref{nullhypothesis:first}}$] \label{nullhypothesis:first}
For all news articles, each statistic feature calculated out of the surprise scores from the true LM has no statistically significant difference with the corresponding one from the satire LM pairwisely.
\end{nullhypothesis}

Here we use the Wilcoxon signed-rank test \cite{mcdonald2009handbook}, which is often used to determine whether two related samples have the same distribution or not. By applying this test on every pair of statistic features, we obtained the p-value $\ll 0.001$ from both satire and true news sample test pairs, which means the Null Hypothesis $H_{\ref{nullhypothesis:first}}$ is rejected and there is a significant difference between true/satire surprise score statistics from both true and satire LMs.

\subsection{Implementation Details}
In this work, we implement the pipeline using typical neural network modules and mediocre settings in each part, comparing to other classification methods with complex network structures or delicate embeddings.
For the language model part, the size of the word embeddings of the encoder is 200. The RNN module is a typical 2-layer LSTM with 200 hidden units per layer. A dropout rate of 0.2 is applied when training the language model. Both of the true news and satirical news LM are trained with 6 epochs.
For the classifier, a typical SVM\footnote{\href{https://scikit-learn.org/stable/modules/generated/sklearn.svm.SVC.html}{sklearn.svm.SVC}} with linear/polynomial kernel is used. 
\begin{figure*}[t]
\centering
  \includegraphics[width=\textwidth]{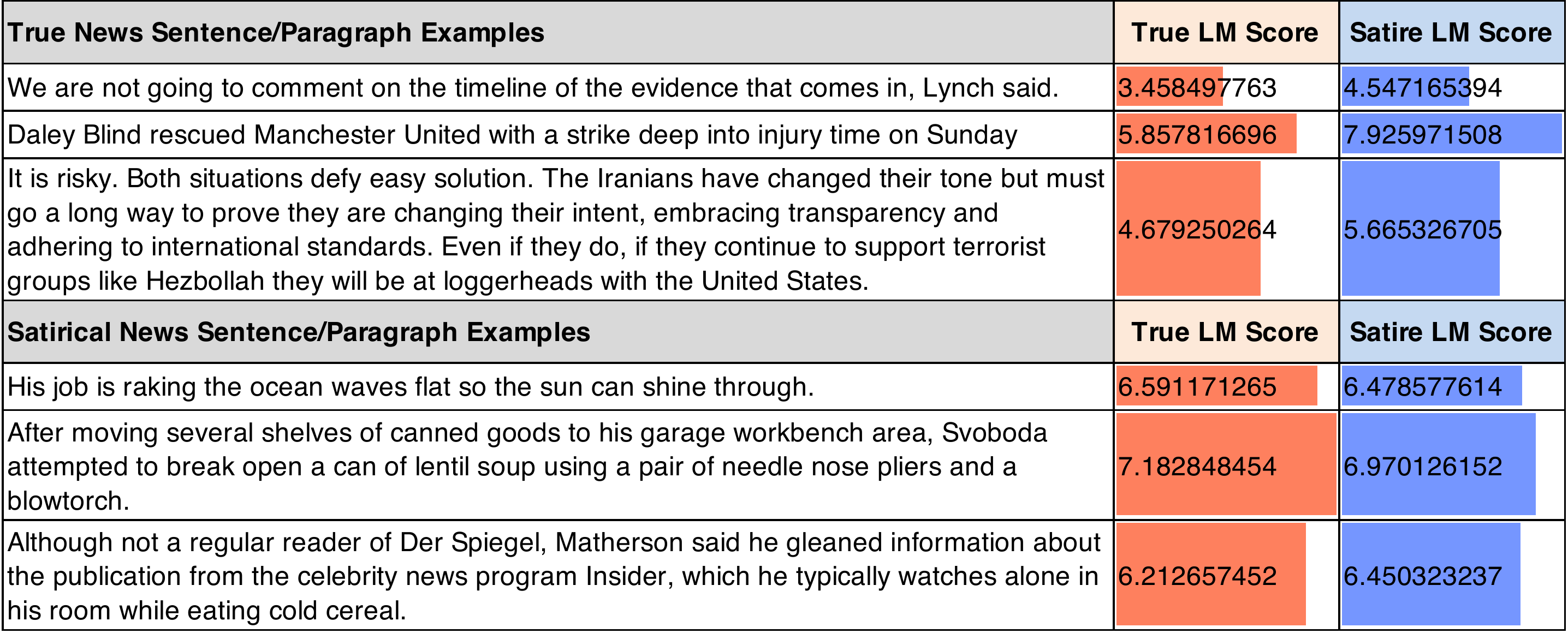}
  \caption{Some examples of some selected news sentences/paragraphs with their surprise scores specified by the true news LM and the satirical news LM.}
  \label{fig:examples}
  \vspace{-10pt}
\end{figure*}

\subsection{Evaluation Results and Analysis}
\begin{figure}[]
  \centering
  \begin{minipage}[t]{0.45\textwidth}
    \includegraphics[width=\textwidth]{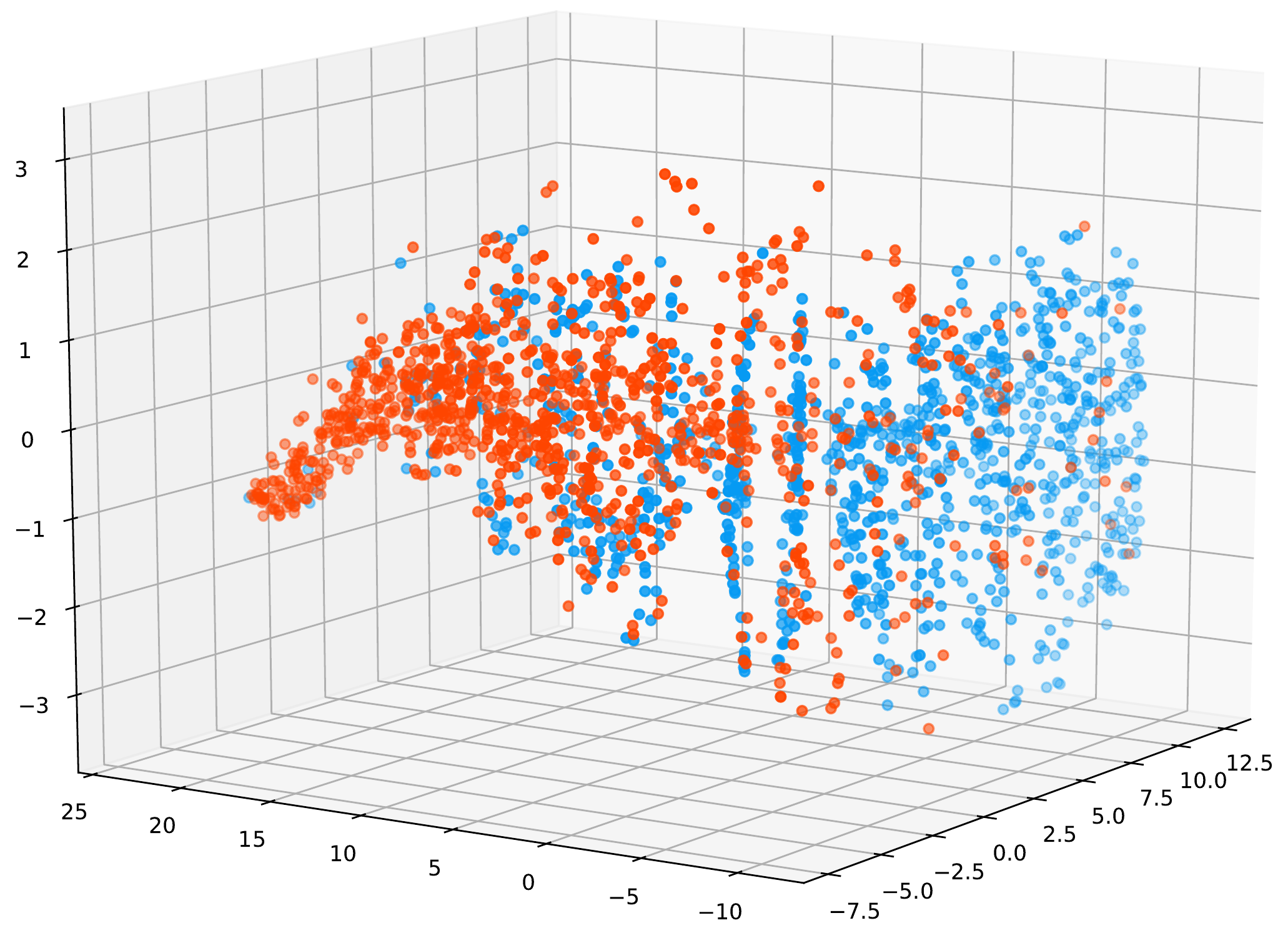}
    \caption{A t-SNE visualization of 1000 randomly sampled feature vectors of true news (red)/satirical news (blue) from Train SVM part of data.}
    \label{fig:tsne}
  \end{minipage}
  \hfill
  \begin{minipage}[t]{0.45\textwidth}
    \includegraphics[width=\textwidth]{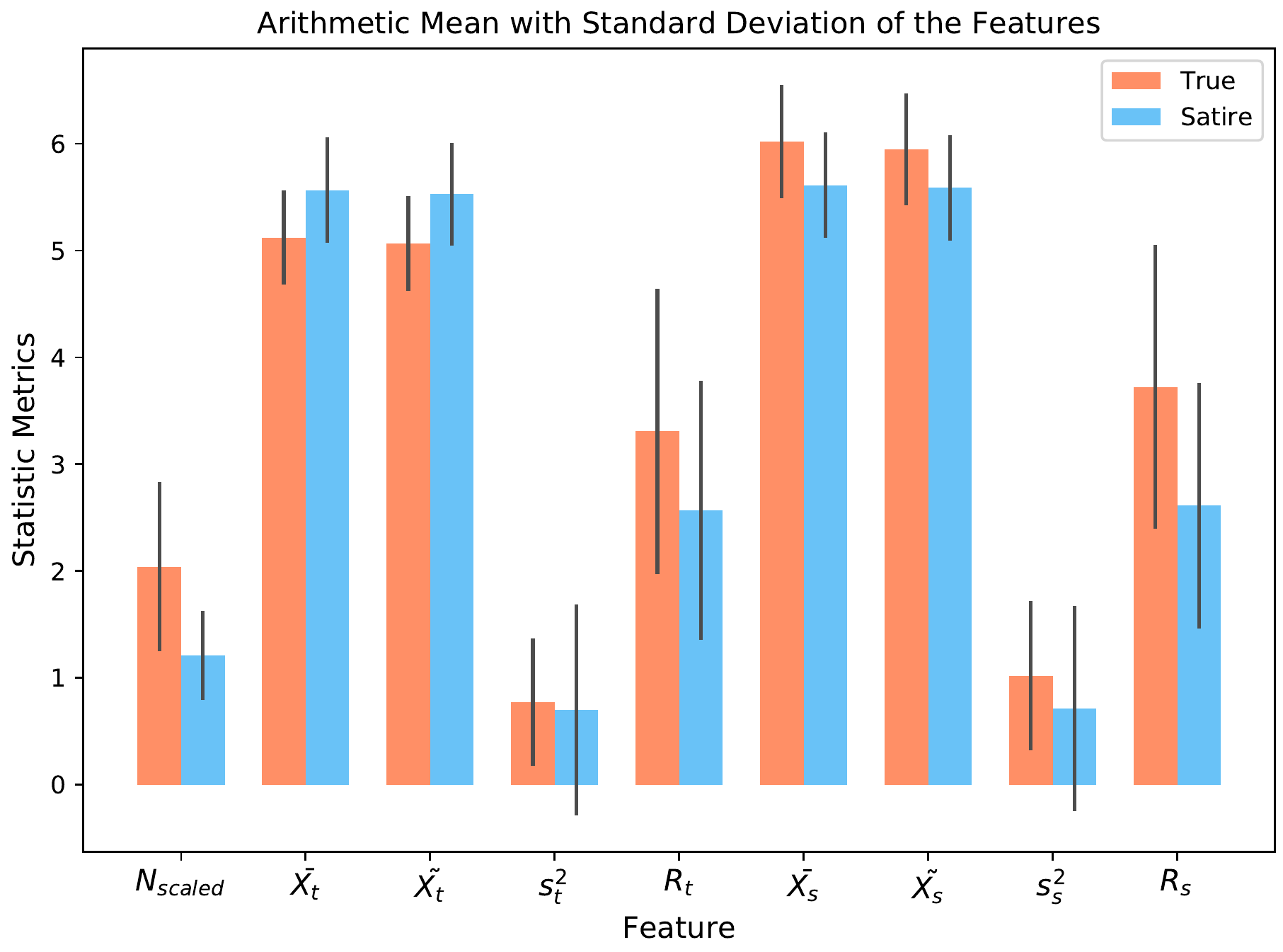}
    \caption{An illustration of the comparison of avg. and std on each feature between true news (red) and satirical (blue). Feature $N$ is scaled by $0.1$.}
    \label{fig:avg}
  \end{minipage}
  \vspace{-10pt}
\end{figure}

Two different language models of true/satirical news giving surprise scores to each sentences, which finally forms a feature vector for one piece of news. Figure \ref{fig:examples} shows some example sentences with their surprise scores on two kinds of LMs. For the true news samples, the surprise scores are generally lower than the satirical news samples. Meanwhile, the scores from true news LM are lower than they are from the satire LM, which confirmed our hypothesis. For the satirical news samples, with their characteristics, the scores not only rise higher but become difficult to distinguish on both sides. This is also abstractly reflected in Figure \ref{fig:avg}.

The statistics calculated from the score sequences depict the different behavior on different LMs from a higher perspective. Figure \ref{fig:tsne} shows that by using the proposed feature vector, it is clear to see the separation of true news samples and satirical news samples even in 3D t-SNE illustration. Table \ref{tab:features} providing a further evidence: with features incrementing, the classification performance improves accordingly. Meanwhile, it also reflects the feature importance when controlling different selection of features. Here we report the experimental results on both validation dataset and test dataset, in order to present the upper bound as well as the generalization performance of this method. The results shows a coherent behavior on both dataset as the number of feature/feature-pairs increments.
The features generated using this method can reflect some statistical differences between two kinds of news data. Figure \ref{fig:avg} is an illustration from a macro perspective using the arithmetic mean for each feature. Visible difference on these two series of data could be found from the histogram plot.

\begin{table}[t]
\centering
\caption{Performance results with increasing number of feature/feature-pairs. 1: Mean, 2: Mean + Median, 3: Mean + Median + Sample Variance, 4: Mean + Median + Sample Variance + Range, 5: Mean + Median + Sample Variance + Range + Sample Size.}
\label{tab:features}
\resizebox{0.75\textwidth}{!}{%
\begin{tabular}{|l|l|l|l|l||l|l|l|l|l|}
\hline
Validation & Acc   & Pre   & Rec   & F1    & Test$\quad$ & Acc   & Pre   & Rec   & F1    \\ \hline
1          & 96.35 & 92.01 & 82.83 & 86.74 & 1    & 94.82 & 90.51 & 77.37 & 82.35 \\ \hline
2          & 96.39 & 92.08 & 83.09 & 86.93 & 2    & 94.76 & 90.45 & 77.04 & 82.08 \\ \hline
3          & 96.90 & 92.61 & 86.32 & 89.16 & 3    & 95.23 & 91.32 & 79.35 & 84.06 \\ \hline
4          & 97.38 & 93.17 & 89.26 & 91.10  & 4    & 96.06 & 92.23 & 83.92 & 87.50  \\ \hline
5          & \textbf{97.97} & \textbf{94.55} & \textbf{92.00}    & \textbf{93.23} & 5    & \textbf{96.82} & \textbf{93.67} & \textbf{87.34} & \textbf{90.19} \\ \hline
\end{tabular}%
}
\vspace{-10pt}
\end{table}

A further consideration is concerning the level of importance of each investigated feature. Here we look into a uni-variate feature selection metric mutual information (MI) \cite{kraskov2004estimating}, which is described in \emph{Eq. 2.28} of \cite{cover2006elements}:
\begin{eqnarray}
I(X;Y) = \sum\limits_{y \in \mathcal{Y}}\sum\limits_{x \in \mathcal{X}} p(x,y) \,\textrm{log}\left(\frac{p(x,y)}{p(x)p(y)}\right) 
\end{eqnarray}
where feature $X$ and target $Y$ are discrete random variables. The MI score here depicts how much the uncertainty of the classification is eliminated when given feature X. We calculate the MI scores between each of the features and the classification target. Thus the higher the MI score is, the more contribution of the corresponding feature will make in classification.

\begin{table*}[t]
\centering
\caption{Experimental result comparison of four different methods on satirical news detection task in \emph{Accuracy}, \emph{Precision}, \emph{Recall}, and \emph{F1 Score}. Results being compared are originally listed in the work by Yang et al. \cite{yang2017satirical} and De Sarkar et al. \cite{de2018attending}.}
\begin{tabular}{|l||l|l|l|l||l|l|l|l|}
\hline
                                 & \multicolumn{4}{c||}{Validation}                                   & \multicolumn{4}{c|}{Test}                                        \\ \hline
Method                                 & Acc            & Pre            & Rec            & F1             & Acc            & Pre            & Rec           & F1             \\ \hline
Rubin et al. \cite{rubin2016fake} & 97.73          & 90.21          & 81.92          & 85.86          & 97.79          & 93.47          & 82.95         & 87.90           \\ \hline
Yang et al. \cite{yang2017satirical}                             & \textbf{98.54} & 93.31          & 89.01          & 91.11          & \textbf{98.39} & 93.51 & \textbf{89.5} & \textbf{91.46} \\ \hline
De Sarkar et al. \cite{de2018attending}                            & 98.18          & 94.15          & 86.55          & 90.19          & 98.31          & 93.45          & 86.01         & 89.57          \\ \hline\hline
This work SVM-Linear                       & 97.93          & 94.46 & 91.79 & 93.07 & 96.67          & 93.41          & 86.62         & 89.65          \\ \hline
This work SVM-Poly                        & 97.97          & \textbf{94.55} & \textbf{92.00} & \textbf{93.23} & 96.82          & \textbf{93.67}          & 87.34         & 90.19          \\ \hline
\end{tabular}

\label{tab:experimentalresult}
\vspace{-10pt}
\end{table*}

\begin{figure*}[h!]
\vspace{-10pt}
  \includegraphics[width=\linewidth]{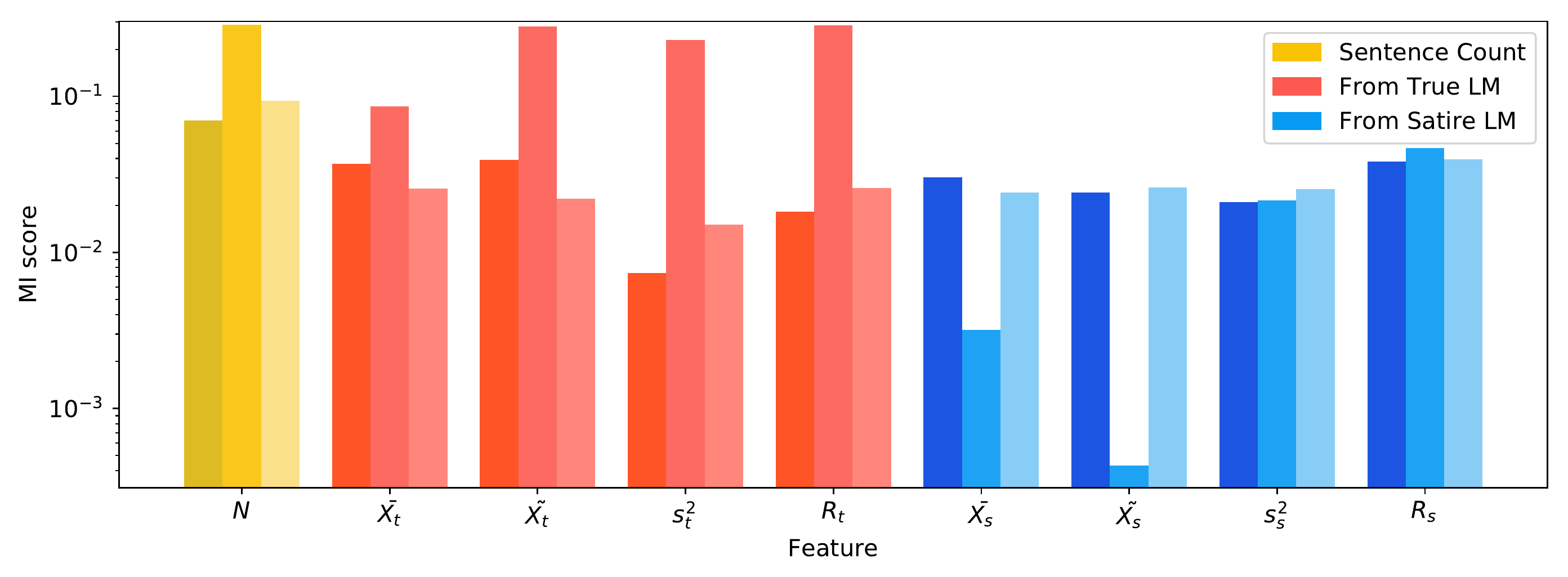}
  \caption{An illustration of Mutual Information feature analysis on the feature sample size, which is sentence (or paragraph) count $N$ (yellow) and paired features from true news LM (red)/satirical news LM (blue). Each feature of Train/Validation/Test data appears from left to right accordingly across each group of the histograms. $Y$-axis is log-scaled.}
  \label{fig:featureanalysis}
  \vspace{-10pt}
\end{figure*}

The MI scores are illustrated in pairs in Figure \ref{fig:featureanalysis} for the training LM data, validation data, and test data. By interpreting the scores, we found that features such as $N$ and all of the features from true LM are of significant importance on validation data, and ${R_s}$ for all dataset play a great role in making decisions in determining news category, while features such as $\bar{X_s}$ and $\tilde{X_s}$ of validation data are obviously of less utility for classification. Although as shown in Figure \ref{fig:avg} there is a less significant difference in the feature pair sample variance, our model has the potential to distinguish and utilize these features. Therefore, the contribution of each feature varies in the classification on different dataset. Furthermore, there is also a visible complementarity shown on each feature item in pairs: if one feature from true news LM has a low MI score, its paired corresponding feature will raise.

As shown in Table \ref{tab:experimentalresult}, our method outperforms the other three methods in Rubin et al. \cite{rubin2016fake}, Yang et al. \cite{yang2017satirical} and De Sarkar et al. \cite{de2018attending} precision, recall, and F1 score on the validation dataset and achieves competitive results on the test dataset. Upon further investigation of the corpora we use, we found that one of the satirical news source proposed by Yang et al. \cite{yang2017satirical} and De Sarkar et al. \cite{de2018attending}, \emph{Ossurworld}\footnote{\href{https://ossurworld.com/}{https://ossurworld.com/}} from the test dataset, is questionable to be categorized and utilized as `satirical news'. This website is a blog of \emph{Irreverence, Irony, Insouciance \& Iconoclasm} as mentioned on their headline instead of news. Although it truly focuses on irony and publishes some satire content and fake news, a considerable number of blog posts such as ironic film reviews are neither satire nor in the form of news. Therefore, it is likely some of this irrelevance in data results in a negative impact on the performance of our method. 

Moreover, our method is not influenced by the potential problems mentioned by McHardy et al. \cite{mchardy2019adversarial} in Section \ref{RelatedWork}, because the language models outputs, surprise scores, are just numerical values, which will not contain any semantic information or learn any fine-grained details like the models \cite{yang2017satirical} and \cite{de2018attending} possibly did. Also, the satire training and testing data are from diverse sources. For the satire part, the LM training and classifier data are from \emph{Onion} and \emph{the Spoof}. The validation data and testing data are from many satire sources listed in Section \ref{ExperimentandEvaluation}. By using data from different news sources, the objectivity of the data and the method can be mutually guaranteed.

\section{Conclusion}
Inspired by the idiom `Birds of a feather flock together', we proposed a new method for satirical news classification that leverages language model output distribution divergence. By leveraging the surprise scores from different language models, the satirical news was differentiated from true news articles effectively. This method is not only free from extracting numerous linguistic features as previous works did, but also it does not require any sophisticated neural network structures or advanced embeddings. More importantly, this proposed method proves the value of the selected statistical features from a language model output, and shows the effectiveness of these features in depicting the characteristics of the corresponding document category.

This  work  is  supported  in  part  by  the U.S. NSF grants 1838147 and 1838145.  We  also  thank  anonymous reviewers for their helpful feedback.
%
%
%
\bibliographystyle{splncs04}
\bibliography{mybib}

\begin{thebibliography}{10}
\providecommand{\url}[1]{\texttt{#1}}
\providecommand{\urlprefix}{URL }
\providecommand{\doi}[1]{https://doi.org/#1}

\bibitem{burfoot2009automatic}
Burfoot, C., Baldwin, T.: Automatic satire detection: Are you having a laugh?
  In: Proceedings of the ACL-IJCNLP 2009 conference short papers. pp. 161--164
  (2009)

\bibitem{cover2006elements}
Cover, T.M., Thomas, J.A.: Elements of information theory (2006)

\bibitem{de2018attending}
De~Sarkar, S., Yang, F., Mukherjee, A.: Attending sentences to detect satirical
  fake news. In: Proceedings of COLING 2018. pp. 3371--3380 (2018)

\bibitem{DragutYSM10}
Dragut, E.C., Yu, C.T., Sistla, A.P., Meng, W.: Construction of a sentimental
  word dictionary. In: CIKM. pp. 1761--1764 (2010)

\bibitem{DBLP:journals/widm/HeHMOD20}
He, L., Han, C., Mukherjee, A., Obradovic, Z., Dragut, E.C.: On the dynamics of
  user engagement in news comment media. Wiley Interdiscip. Rev. Data Min.
  Knowl. Discov.  \textbf{10}(1) (2020)

\bibitem{kraskov2004estimating}
Kraskov, A., St{\"o}gbauer, H., Grassberger, P.: Estimating mutual information.
  Physical review E  \textbf{69}(6),  066138 (2004)

\bibitem{mcdonald2009handbook}
McDonald, J.H.: Handbook of biological statistics, vol.~2 (2009)

\bibitem{mchardy2019adversarial}
McHardy, R., Adel, H., Klinger, R.: Adversarial training for satire detection:
  Controlling for confounding variables. In: Proceedings of NAACL-HLT 2019. pp.
  660--665 (2019)

\bibitem{mikolov2010recurrent}
Mikolov, T., Karafi{\'a}t, M., Burget, L., {\v{C}}ernock{\`y}, J., Khudanpur,
  S.: Recurrent neural network based language model. In: Proceedings of
  Interspeech 2010 (2010)

\bibitem{reyes2012humor}
Reyes, A., Rosso, P., Buscaldi, D.: From humor recognition to irony detection:
  The figurative language of social media. Data \& Knowledge Engineering
  \textbf{74},  1--12 (2012)

\bibitem{rubin2016fake}
Rubin, V., Conroy, N., Chen, Y., Cornwell, S.: Fake news or truth? using
  satirical cues to detect potentially misleading news. In: Proceedings of the
  second workshop on computational approaches to deception detection. pp. 7--17
  (2016)

\bibitem{rubin2015deception}
Rubin, V.L., Chen, Y., Conroy, N.J.: Deception detection for news: three types
  of fakes. In: Proceedings of the 78th ASIS\&T Annual Meeting: Information
  Science with Impact: Research in and for the Community. p.~83. American
  Society for Information Science (2015)

\bibitem{SchneiderD15}
Schneider, A.T., Dragut, E.C.: Towards debugging sentiment lexicons. In: ACL.
  pp. 1024--1034 (2015)

\bibitem{StanojevicADO19}
Stanojevic, M., Alshehri, J., Dragut, E.C., Obradovic, Z.: Biased news data
  influence on classifying social media posts. In: NEwsIR@SIGIR. vol.~2411,
  pp.~3--8 (2019)

\bibitem{trinh2018simple}
Trinh, T.H., Le, Q.V.: A simple method for commonsense reasoning. arXiv
  preprint arXiv:1806.02847  (2018)

\bibitem{yang2017satirical}
Yang, F., Mukherjee, A., Dragut, E.: Satirical news detection and analysis
  using attention mechanism and linguistic features. In: Proceedings of EMNLP
  2017. pp. 1979--1989 (2017)

\end{thebibliography}
\end{document}